\documentclass[journal]{IEEEtran}
\usepackage{graphicx,amssymb,epsfig}
\usepackage{amsmath,epsfig}
\usepackage{cite,subfig}

\newcommand{\mt}{\mathbb}   
\newcommand{\mc}{\mathcal}  

\begin{document}
\title{Spatially-Adaptive Reconstruction in Computed Tomography Based on Statistical Learning}
\author{Joseph Shtok, Michael Zibulevsky, Member, IEEE, \\
         and Michael Elad, Senior Member, IEEE.
        \thanks{All authors are with the Computer Science Department,
        Technion - Israel Institute of Technology, Israel.}
        \thanks
        {This research was supported by the European
        Community's FP7-FET program, SMALL project, under grant agreement
        no. 225913.}
        \thanks{This research was supported by Gurwin Family Fund.}
} \maketitle

\begin{abstract}
We propose a direct reconstruction algorithm for Computed
Tomography, based on a local fusion of a few preliminary
image estimates by means of a non-linear fusion rule. One such
rule is based on a signal denoising technique which is spatially
adaptive to the unknown local smoothness. Another, more powerful
fusion rule, is based on a neural network trained off-line with a
high-quality training set of images. Two types of linear reconstruction
algorithms for the preliminary images are employed for two different
reconstruction tasks. For an entire image reconstruction from full
projection data, the proposed scheme uses a sequence of Filtered
Back-Projection algorithms with a gradually growing cut-off
frequency. To recover a Region Of Interest only from local
projections, statistically-trained linear reconstruction algorithms
are employed. Numerical experiments display the improvement in
reconstruction quality when compared to linear reconstruction
algorithms.
\end{abstract}

\section{Introduction}

\IEEEPARstart{T}{he Filtered} Back-Projection (FBP) algorithm is
extensively used for image reconstruction in the Computed Tomography
(CT). This algorithm implements in the 2-D case a discretization of the inverse
Radon transform. Despite the popularity of this
method, its drawbacks are non-negligible: FBP fails to account for
the numerous physical phenomena present in the data acquisition
process (resulting in the well-known streak artifacts) and suffers
from discretization errors. Moreover, it lacks the flexibility
required to process partial input data, like in the case of
projections truncated to a Region Of Interest (ROI), or in the
case where projections are restricted to a limited angle range.

Numerous techniques were developed to improve the performance of
FBP. Many of them modify the filters, applied to the projection
data. The basic problem with the standard Ram-Lak filter \cite{RaLa71}
is the high-frequency noise amplification. It is commonly treated by using an additional
low-pass filter, which cut-off frequency is compatible with the
expected bandwidth of the signal. Such an approach requires tuning the cut-off
frequency and other parameters of the low-pass filter. The work reported
in \cite{KiKi96}, for instance, is dedicated to tuning these
parameters for lesion detectability.

One important task in the clinical CT reconstruction is to recover
the CT image in a Region Of Interest in the patient's body using
low-exposure scan. To that end, there exist techniques which require
projections data only in the neighborhood of the ROI, in addition to a
small number of full projections. These techniques employ an FBP
modified to compute wavelet coefficients of the sinogram or the sought image by
applying wavelet ramp filters \cite{OlDe94,DeBr95, FaWa97}. This
allows to recover the high spatial frequencies in the ROI from local
data. The low frequencies require full projections; however, since they
demand much smaller angular sampling rate,
a local reconstruction can be done with much less
X-ray exposure. There are, however, no proposed modifications of FBP
that undertake the ROI reconstruction in the setup where only
projections in the neighborhood of the ROI are measured. Such
algorithm would allow further reduction in X-ray dosage, reduce scan
duration and dismiss the necessity for the registration of the
projection data.

To improve the reconstruction quality and overcome the above
drawbacks, statistically-based iterative algorithms were developed.
They use an elaborate model for the CT scan measurements, including
sources of noise and partial projections data. We refer to \cite{ElFe02} for
a detailed example of such an algorithm and to \cite{Fess06} for a
broad overview of iterative methods. Unfortunately, high computational
cost of these algorithms restricts their use in clinical CT
scanners.

Superiority of the statistically-based methods stems from the fact
the image is reconstructed in a non-linear, locally-adaptive manner,
relying only on the available projections data. For instance, with the
method described in \cite{ElFe02} the reconstruction is performed
via optimization of a Penalized Weighted Least Squares (PWLS), with
a penalty promoting local smoothness in the image domain in an
edge-preserving manner. Such behavior can not be achieved with a
linear, spatially-invariant algorithm like FBP. Therefore, an
algorithm combining the advantages of both approaches is desired: a
direct (and, therefore, fast) processing of the available
data, on one hand, with a non-linear, locally-adaptive nature on the
other hand.

One such algorithm is developed in \cite{AnSa02}. The proposed
algorithm uses a powerful filtering technique, involving a training
procedure. It employs an exemplar-based classification of the
sinogram data patches, combined with training of local 2-D projection
filters, which results in a non-linear overall filtering procedure.

In this work we propose a non-linear, locally-adaptive reconstruction
scheme, based on example-based statistical learning of its
components. The scheme consists of two stages: first, a sequence of
linear FBP-like transforms are applied to the available projections
data, resulting in a number of preliminary image estimates. Then, a
local non-linear fusion of these estimates is performed to produce
the final image.

In the setup of an entire image reconstruction from a full-scan, the
linear estimates are obtained with the FBP algorithm with a varying
cut-off frequency of the projections filter. A more important goal is
to reconstruct an ROI from truncated projections. Since FBP does not
perform well in the absence of global projections, we have developed a
more flexible linear reconstruction scheme called AFBP\cite{ShEl08}. It
generalizes the FBP and employs more powerful filters in the sinogram
and image domains. The convolution kernels for these filters are
derived via statistical training which accounts for missing data and the
desired reconstruction properties. In the proposed algorithm, a set
of linear estimates of the ROI image are computed by a number of AFBP
versions only from projections through a disk containing the ROI,
which radius is $110\%$ of the ROI radius. Then the ROI is recovered by a
neural network from these preliminary reconstructions by a non-linear,
learned fusion rule.

While the computational cost of this procedure is only about ten
times the cost of the FBP algorithm\footnote{In practice, the run
time can be reduced to one FBP computation by parallel
execution on small number of cores.}, two main features distinguish the
proposed method from any spatially-invariant reconstruction
transform. First, the components of our scheme are designed to work
with partial data (truncated projections). Second, the reconstruction
is locally adaptive in the image domain, which allows to reduce the
noise present in linear estimates and to preserve edges and texture
in a better way.  The proposed method is labeled as SPADES (SPatially
ADaptive EStimator).

This paper extends our work presented earlier in conference publications
\cite{ShEl08,ShEl09}. Most of the presented material is new,
except for the description of the linear AFBP scheme. The SPADES
algorithm, which is also briefly presented in \cite{ShEl09}, has been
revised and improved.

The paper begins with Section \ref{sect:Model}, containing preliminaries
and notation. For a theoretical motivation of
SPADES we present the locally-adaptive denoising algorithm of Lepski,
Goldenschluger and Nemirovsky \cite{GoNe97} and extend it to CT reconstruction
setup (Section \ref{sect:LeGoNe}). Then SPADES is described and demonstrated
in Section \ref{sect:SPADES_FBP}. The AFBP scheme is
introduced in Section \ref{sect:AFBP}. Then, the SPADES
is extended to ROI setup in Section \ref{sect:SPADES_AFBP}. Discussion follows
in Section \ref{sect:EPILOG}.

\section{A model of 2-D Transmission Tomography} \label{sect:Model}
A 2-D slice of a physical object is represented by the attenuation map $f(x)$,
defined in the domain $\mc{A}\subset\mt{R}^2$ -- this is the image
recovered by the CT reconstruction. This map, assumed to have a
support radius $R$, is projected along straight lines by means of
the 2-D Radon transform: for $s\in[-R,R]$,$\theta\in[0,\pi]$, the
transform $g =\mathbf{R}f$ is defined by
\begin{equation}
g_{\theta}(s)=\int_{t\in \mt{R}}f(s\cdot cos(\theta)-t\cdot
sin(\theta), s\cdot sin(\theta)+t\cdot cos(\theta))dt.
\end{equation}
We denote the range of $\mathbf{R}$ by $\mc{P}\subset \mt{R}^2$. The
adjoint transform $\mathbf{R}^*$, also known as a Back-Projection, is
defined by
\begin{equation}
(\mathbf{R}^*g)(x)=\int_{\theta}g_{\theta}([cos(\theta),sin(\theta)]\cdot
x)d\theta.
\end{equation}
In the discrete setting, the Radon transform of $f(x)$ is sampled at
a large number of fixed angles (evenly covering the range $[0,\pi]$)
and fixed signed distances $s$ (bins). The matrix image $f(x)$ is
computed from these samples by a discrete reconstruction algorithm.

Let $\ell=\ell(\theta,s)$ be a line which makes the angle
$\pi/2-\theta$ with the $x$ axis and passes at a distance $s$ from
the origin. To each such line there corresponds a detector which
counts the number $y_{\ell}$ of photons in a specified time
interval during the scan.  Due to the limited photon count, the values
$y_{\ell}$ are modeled as realizations of random variables:
$y_{\ell}\sim
\mbox{Poisson}\left(I_{0}e^{-(\mathbf{R}f)_{\ell}}\right)$. The
X-ray source intensity $I_0$ determines the parameters of the
Poisson distribution, thus controlling the noise level. The Maximum
Likelihood (ML) estimate of $\mathbf{R}f$ from the measurements
$y_{\ell}$ is $g_{\theta}(s)=-log (y_{\ell}/I_0)$. In the ideal case
where $y_{\ell}=\mt{E}(Y_{\ell})=I_0e^{-(\mathbf{R}f)_{\ell}}$ (not
attainable in reality since the expectation of $Y_{\ell}$ is not, in
general, an integer), ML estimate is the true Radon transform of
$f(x)$.

More realistic modeling of the CT scan takes into account additional
disruptions, such as electronic noise (additive constant in the
Poisson parameter $\lambda_{\ell}$), scatter of X-rays (Compton
effect), crosstalk among the detectors, and more \cite{RiBi06}.

The basic FBP algorithm is defined by means of the
linear reconstruction operator
$\mathbf{T}_{\text{FBP}}=\mathbf{R}^*\circ\mathbf{F}_{\text{Ram-Lak}}$.
Here the filter $\mathbf{F}_{\text{Ram-Lak}}$ is the 1-D convolution kernel,
applied to each projection ( acolumn in the sinogram matrix). The
Ram-Lak kernel $\zeta$ is defined in the Fourier domain by
$\hat{\zeta}(\omega)=|\omega|$. As mentioned earlier, the Ram-Lak
kernel is often smoothed by some low-pass filter compatible to the
reconstructed images and the noise level.

%
%
%
%
%
%
%

\section{From Adaptive Denoising Technique to a Fusion Rule for CT
Reconstruction}\label{sect:LeGoNe}

Our goal is to bridge the gap between the linear and the iterative
algorithms by a direct reconstruction scheme, locally adaptive to the
data. To that end, we employ the technique of filtering with adaptive
kernels, developed originally for signal reconstruction from noisy
measurements. In the classical setup of the problem, a signal $f$ is measured
through
$$
y_i = f(x_i)+\xi_i
$$
where the set $\{x_i\}$ is a sampling grid and $\xi_i$ are
independent normal random variables. The task is to compute an
estimate $\tilde{f}(x)$ which minimizes the $L_2$ norm of the error
$\epsilon=|f - \tilde{f}|$ in an interval of interest.

One basic denoising technique is to apply a linear kernel estimator
$\tilde{f}_{\kappa}=y*\kappa$, where $\kappa$ is a constant convolution kernel
representing a low-pass filter. Such an estimator will mistreat those
regions where the bandwidth of the signal does not match the
bandwidth of the filter (i.e., either the edges will be blurred or
the smooth regions will remain noisy).

A substantial improvement can be achieved by using, at each image
location, a low-pass filter which cut-off frequency matches the
local spatial smoothness of the signal. Thus, in smooth regions a
stronger blur will be used, averaging the noisy values, and near
edges almost no blur will be applied in order not to smear them. Of
course, such knowledge of the local smoothness of the underlying
clean signal is not available, but it can be evaluated using local
image statistics. Therefore, given a sequence of filter kernels with
gradually growing measure of the blur and a good decision rule,
choosing an appropriate filter for each spatial location, a
spatially adaptive signal estimator can be implemented.

An analogue of image filtering with kernel $\kappa$ in CT
reconstruction is a linear reconstruction transform $\mathbf{T}$
with the property that the Point Spread Function (PSF) of the
(shift-invariant) operator $\mathbf{TR}$ is the kernel $\kappa$. It
is easy to obtain a sequence $\mathbf{T}_i,i=1,...,I$ of linear
reconstructors with gradually growing spread of the PSF; therefore
an analogue of spatially-adaptive denoising can be designed for CT
reconstruction, as we detail in Section \ref{LeGoNe-CT}. We should stress
that the algorithm, described and demonstrated therein, is not intended for
practical use: it is a mid-step between the theoretically-based denoising technique
for Gaussian noise and the non-linear CT reconstruction algorithm based on a learning machine.

%
%
%
%
\subsection{Lepski-Goldenshluger-Nemirovsky Estimator}
A locally adaptive estimator for signal denoising was devised in the work
\cite{GoNe97} by A. Goldenshluger and A. Nemirovsky, based on a general
scheme by Lepski. Consider again
the task of recovering a 1-D signal $f(x)$ from noisy observations
$y(x) = f(x)+\xi(x)$ on a discrete grid points $\{x\in \Gamma\}$,
where $\{\xi(x)\}_{x\in \Gamma}$ is a sequence of independent normal
random variables. To estimate $f(x)$ at some point $x_0$, a linear
combination of the neighboring samples $y(x)$ is taken (as is done
by a convolution kernel with corresponding coefficients). The basic
idea of this estimator is demonstrated for simple rectangular windows
centered in $x_0$: $\tilde{f}(x_0)=mean\{y(x)\;|x\in\ \triangle\}$,
where $\triangle =[x_0-\delta,x_0+\delta]$. In practice, order-$m$
Least Squares (LS) polynomial is fitted locally into the data.

The error $|\tilde{f}_{\triangle}(x_0) - f(x_0)|$ consists of the
deterministic {\em dynamic error}
$\omega_f(x_0,\Delta)=mean_{x\in\triangle}|f(x)-f(x_0)|$ and the
{\em stochastic error}
$\zeta(\triangle)=\frac{1}{|\Delta|}\sum_{x\in \triangle}\xi(x)$,
introduced by the noise. As the window $\triangle$ grows, the
dynamic error increases (with the rate related to the signal
smoothness) and the stochastic error decreases, since the noise is
averaged over a larger interval. The goal is to find an optimal window
width, which balances the error components.

In order to approximate the optimal width, the authors of \cite{GoNe97}
employ confidence intervals. Consider a sequence
$\triangle_1,...,\triangle_N$ of windows, centered in $x_0$, with a
growing width. For each $\triangle_i$ there is an estimated upper
bound $\rho_i$ of the stochastic error in this window. For $i=1$ it is
assumed the the window is small enough so that the dynamic error is
dominated by the stochastic one, hence the overall error is bounded
from above by $2\rho_1$. The confidence interval related to
$\triangle_i$ is then defined by
\begin{equation}\label{eq:conf_int}
D_i=[\tilde{f}_{\triangle i}(x_0)-2\rho_i, \tilde{f}_{\triangle
i}(x_0)+2\rho_i].
\end{equation}
Notice that for every index $i$ satisfying
$\omega_f(x_0,\Delta_i)<\zeta(\triangle_i)$, the total error
$|\tilde{f}_{\triangle_i}(x_0) - f(x_0)|$ is smaller than $2\rho_i$
(with the aforementioned high probability). Therefore, the interval
$D_i$ contains the true value $f(x_0)$.

In order to estimate the maximal index $i^*$ with this property, the
intersection of intervals $D_i$ is considered. A simple argument,
presented in \cite{GoNe97}, shows that if $i^+$ is the maximal index
for which
\begin{eqnarray}\label{eq:swr}
\bigcap_{i<=i^+}D_i\neq \emptyset,
\end{eqnarray}
then the estimate error is bounded by
\begin{eqnarray}
|\tilde{f}_{i^+}(x_0) - f(x_0)|\leq 6\rho_{i^*}
\end{eqnarray}
Since the stochastic error $\rho_{i^*}$ is minimal among possible
errors corresponding to different segments $\triangle_i$, the choice
of $\tilde{f}_{i^+}$ is nearly optimal.

As the simple averaging is replaced with order-$m$ LS approximation,
the efficiency becomes comparable to best denoising techniques. It is
proven in \cite{GoNe97} that such algorithm is near-optimal.

Before passing on to the CT reconstruction, we mention that a similar
algorithm was devised by Lepski, Mammen, and Spokoiny in
\cite{LeMa97}. Both works implement a general scheme of Lepski
\cite{Leps90} and seem to share the same approach.\\
%
%
%
%
\subsection{A Switch Rule for CT Reconstruction}\label{LeGoNe-CT}
The denoising technique, described above, requires a preliminary
sequence of linear signal estimates, obtained by filtering the noisy
signal with a corresponding sequence of convolution kernels
(rectangular windows of growing radius, in this case). In the setup
of CT, we use a sequence of linear transforms
$\{\mathbf{T}_i\}_{i=1}^I$, each is the FBP algorithm involving a
low-pass filter with a different cut-off frequency. Let $g$ denote
the noisy sinogram and $\tilde{f_i}=\mathbf{T}_i(g)$ is the sequence
of image estimates. The the noise, present in the reconstructed
images, is of a complicated, data-dependent nature. The denoising
algorithm requires to know the bound $\rho_i(p)$ on the statistical
error in each image location $p$ of each $\tilde{f_i}$; these bounds
can be computed using noise variance in the image domain.

Estimating the noise variance is difficult, and we do not pursue this
task here. Our goal is to show that {\em given the necessary
information for the switch rule, locally adaptive CT reconstruction
is possible}. Once the evidence for such success is obtained, we
propose a different fusion rule, based on a learning machine.
Therefore, we simulate a large number on noise instances to compute
the variance $\lambda_i(p)$ of the noise at the location $p$ in the
image estimate $\tilde{f}_i$. Then, instead of estimating the bound
$\rho_i$ on the stochastic error in $p$ as it is done in
\cite{GoNe97}, we compute
\begin{equation}
\rho_i(p) = \kappa(\sqrt{\lambda_i(p)})^q
\end{equation}
where the parameters $\kappa,q$ are tuned using a grid search on the
training set. These two parameters are required to calibrate the
numerical values of the data obtained in CT reconstruction process
and are set exactly once for the given setup.

The confidence intervals are computed by the formula
\ref{eq:conf_int}. Then a switch rule, expressed by the condition
(\ref{eq:swr}), is applied to compute the index $i^+$. The output
value in location $p$ is $\tilde{f}_{i^+}(p)$. We label the described
algorithm as {\em LeGoNe} after the authors of the prototypical denoising technique (Lepskii, Goldenshluger, Nemirovski).

%
%
%
%
\subsection{Numerical Experiment - CT reconstruction with LeGoNe}
The quality measure we use is the
Signal-to-Noise Ratio (SNR), defined for the true signal $f_0$
and its estimate $\tilde{f}$ by
\begin{equation}
SNR(f_0,\tilde{f}) =
-20log_{10}\dfrac{\|f_0-\tilde{f}\|_2}{\|f_0\|_2}.
\end{equation}

\begin{figure}[htbp]
\centering
\includegraphics[width=7.5cm]{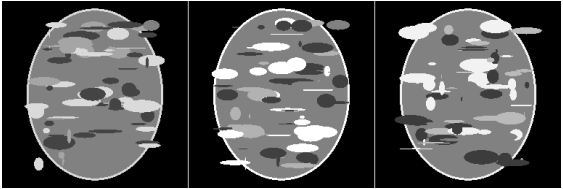}
\caption{Examples of geometrical phantoms used in the experiments.}
\label{fig:RanPhan_samples}
\end{figure}

In this experiment we use a set of randomly generated $256\times 256$
geometric phantoms. Each phantom constitute of a large ellipse with
boundary and a constant background, filled with many smaller ellipses
with randomly chosen centers and radii. There are four intensity
levels in each image (also randomly chosen for each phantom).
Examples of the phantoms are presented in Figure
\ref{fig:RanPhan_samples}.

In our simulations of CT projection and reconstruction, the
projection sets (sinograms) of the reference images are computed
using a pixel-driven implementation of a discrete Radon transform
$\mathbf{R}$. Its adjoint transform is
employed in the reconstruction process. The sinogram noise is
generated in accordance to the statistical model, described in
Section \ref{sect:Model} (see also the Appendix section for the
details).

\begin{figure}[htbp]
\centering
\includegraphics[width=8.5cm]{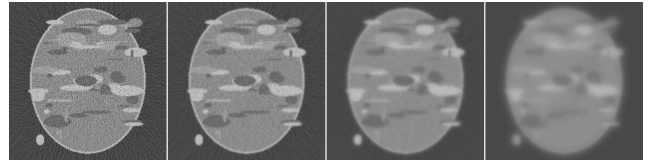}
\caption{Sequence of linear FBP estimates for the projected phantom,
with a growing degree of blur.} \label{fig:Legon_CT_01}
\end{figure}

The sequence of the FBP algorithms
$\{\mathbf{S}_i\}_{i=1}^I$ is generated by applying a Butterworth
window with growing radius to the Ram-Lak convolution kernel $\kappa_0$ in the
Fourier domain:
\begin{equation}\label{eq:BW}
\kappa_i = \kappa_0*\mathbf{F}^{-1}(H_i),\;\;\;H = H(p,q_i) =
\frac{1}{1+q_i^{2p}}
\end{equation}
Here $\mathbf{F}$ is the 1-D Discrete Fourier Transform. The
parameter $p$ controls the steepness of
the window roll-off, and $q$ (cut-off frequency) determines its
width.  $\{q_i\}_{i=1}^I$ is a monotonously decreasing sequence. Figure
\ref{fig:Legon_CT_01} displays (part of) the corresponding sequence of
$\tilde{f_i}=\mathbf{S}_ig$ obtained from a sinogram $g$ of a geometric phantom.

\begin{figure}[htbp]
\centering
\includegraphics[width=9cm]{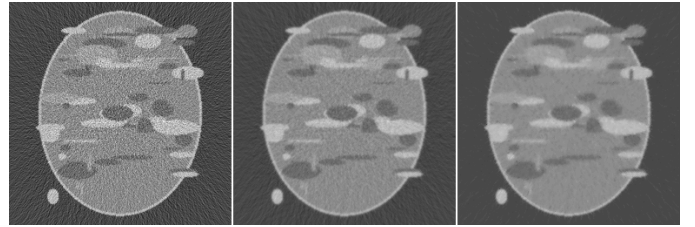}
\caption{Reconstruction results, left to right: FBP with ideal Ram-Lak filter, FBP with optimally apodized Ram-Lak filter (SNR = 15.72 dB), LeGoNe (SNR = 16.13 dB).} \label{fig:Legon_CT_02}
\end{figure}

The LeGoNe reconstruction algorithm, as described earlier, is applied
using the images $\{\tilde{f}_i\}_{i=1}^I$. Comparison of various
reconstruction results can be observed in Figure
\ref{fig:Legon_CT_02}. Notice that, despite the modest increase in
the SNR value, much of the noise present in FBP reconstructions is
removed in the LeGoNe estimate.

%

\begin{figure}[htbp]
\centering
\includegraphics[width=4cm]{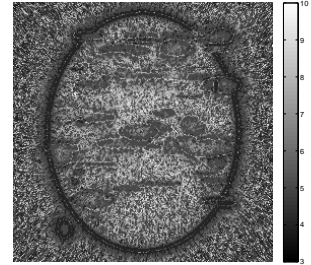}
\caption{The switch map of the LeGoNe algorithm.}
\label{fig:Legon_CT_03}
\end{figure}

In the switch map generated by the LeGoNe (Figure
\ref{fig:Legon_CT_03}), an intensity value in each location $p$ is the
index $i^+$ chosen for the outcome image in $p$. It can be
observed that in smooth areas of the phantoms the algorithm prefers
higher indices (since in blurred images the stochastic error is lower)
and near the edges of the ellipses, lower indices are chosen.

In practice, the LeGoNe algorithm is not our method of choice. First,
the underlying denoising technique is optimal only in the mini-max sense and
up to a constant. Second, the evaluation of noise statistics in the
image domain is difficult. More importantly, we
can not use outputs of other reconstruction algorithms as preliminary
results, to be further improved by LeGoNe.
%
%
%
%
%
%
%

\section{SPADES - Local Fusion Based on a Neural Network}\label{sect:SPADES_FBP}
To build a more powerful local fusion rule we resort to a neural
network. Indeed, it is difficult to devise an analytical rule to
approximate the true image value from the set of preliminary
reconstructions $\tilde{f_i}=\mathbf{T}_ig$. The problem is severed by the
non-homogeneous, data- and algorithm-dependent noise, present in the
image. Yet, by the analogy with the situation discussed in the
previous section, we can hope that the set of values
$\{v_i=\tilde{f_i}(x)\}_{i=1}^I$, for carefully chosen estimates
$\tilde{f_i}$, contains information on $f(x)$ which is more accurate
than any of the individual estimates; in other words, there exists a
function $\Psi = \Psi(v_1,..., v_I)$ that will produce a more
accurate image reconstruction when applied point-wise to the set $\{\tilde{f_i}\}_{i=1}^I$.

A neural network can learn this function via a
training set of reference images $\{f^t\}_{t=1}^T$ and their simulated
estimates $\{\tilde{f}_i^t\}_{1\leq i\leq I,\;1\leq t \leq T}$.
The sinogram data is obtained by
either applying a Radon transform to discrete
reference images, or (in a more realistic setup) by scanning a set of
geometric phantoms in a clinical CT scanner. Then the linear
reconstruction algorithms are applied in order to compute the
training data $\{\tilde{f_i^t}\}$. The learning procedure occurs off-line, and when an unknown object $f^n$ is scanned, the images $\{\tilde{f^n_i}\}_{i=1}^I$ are computed from its projections data and fed to the trained neural network
point-wise, to produce the final outcome.

\begin{figure}[hpbp]
\centering
\includegraphics[width=6.5cm]{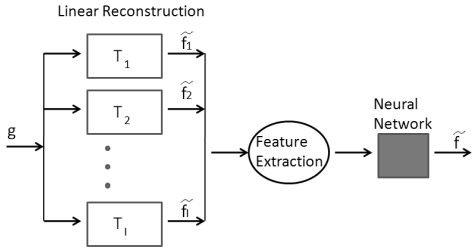}
\caption{SPADES reconstruction scheme}
\label{fig:SPADES_scheme}
\end{figure}

Notice that in this setup we are not confined to use a sequence of linear
reconstruction algorithms with a gradually increasing measure of
blur, like with the LeGoNe technique. Any available preliminary
reconstructions (not necessarily linear) can be fused with the neural network in order to produce a result, expectedly superior to all the participating
versions. The scheme of the resulting SPADES algorithm is given in the Figure
\ref{fig:SPADES_scheme}).

\subsection{Algorithm Description}\label{sect:NN}
\textbf{Definition of the neural network:} We use a single-layer feedforward type neural network. Its output function is defined by
\begin{equation}\label{eq:nneq}
y(x,w,v) = \sum_{j=1}^N v_j\sigma(\sum_{k=1}^K w_{k,j}x_i+w_{K+1,j}),
\end{equation}
where $N$ is the number of neurons, $K$ is the size of the input
vector of features $x$, $w_{k,j}$ is the weight on edge connecting
$k$-th input to $j$-th neuron
and $\sigma(z)=z/(1+|z|)$ is the sigmoid function.

The learning procedure consists in solving the following
optimization problem: given a set of vectors of features
$x^t=[x_1^t,...,x_M^t]$ and the corresponding true values $v(t)$ for $t=1,...,T$, minimize the objective function
\begin{equation}\label{eq:nnobj}
(w^*,v^*) = arg\min_{(w,v)}\{ \sum_{t=1}^T (v(t)- y(x^t,w,v))^2\}
\end{equation}
This function is non-convex. In our experiments it is minimized using the
Matlab routine {\em fminunc.m}. It performs an unconstrained optimization
using the BFGS Quasi-Newton method with a cubic line search procedure.

\noindent\textbf{Training set design:} Notice that intensity values of the reference image $f(p)$ can vary significantly over different regions. We reduce the variability of the data fed to the neural network by using relative values of the images: the vector of
features $x_p$ corresponding to a spatial location $p$ is build as
follows. The first $I$ values are set to
\begin{equation}
x_p(i) =\tilde{f_i}(p)-\bar{f}(p),
\end{equation}
where $\bar{f}$ is the best\footnote{In the SNR sense. We use the training set to determine which linear estimator is the best.} available estimate of $f(x)$. Other entries of $x_p$ consist of image values in a small neighborhood $\mc{N}_p$ of $p$ taken from $\bar{f}$. This provides the neural network with additional local information about the image at the point $x$, allowing for more accurate restoration of the value $f(x)$.

The corresponding true output value $y_p$, provided along with
the vector $x_p$ in the training stage, is set to $y_p = f(p) -
\bar{f}(p)$. In the reconstruction stage, the final outcome image is obtained by \begin{equation}
\tilde{f}_{nn}(p) = y(x_p,v,w)+\bar{f}(p).
\end{equation}

The dynamic range of the input values for the neural network is normalized to $[0,1]$.
%

%
%
%

\subsection{Numerical Experiment - SPADES on a full-scan data}

We repeat the previous experiment in CT reconstruction, when instead of
the LeGoNe switch rule the fusion is performed by a neural network. The
training data is extracted in the way
described earlier in this section. A network of 24 neurons was
trained with a set of $15800$ vectors of features (sampled from ten
images) and then applied to a test phantom. Ten preliminary versions
of each image were built, using an FBP algorithm with varying degree
of blur.

\begin{figure}[hpbp]
\centering
\includegraphics[width=9cm]{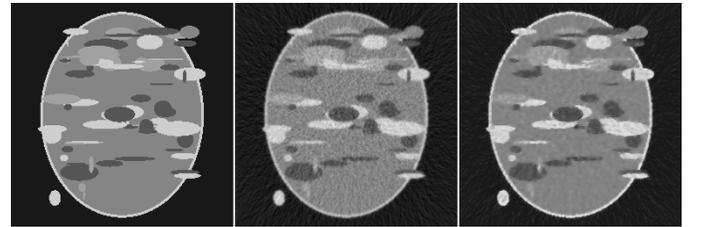}
\caption{Left to right: phantom, best linear reconstruction (SNR =
15.72 dB), neural network output (SNR = 18.00 dB) }
\label{fig:FBP_NN_CT_02}
\end{figure}

The outputs of the FBP and the SPADES algorithms are displayed in Figure
\ref{fig:FBP_NN_CT_02}. It can be observed that the noise streaks,
characteristic for FBP reconstruction, do not appear in the SPADES
output. It is much closer to the piecewise constant
phantom, and the SNR value reflects the quality improvement: it is $2.3$ dB
higher than the linear reconstruction and $1.8$ dB higher than the
LeGoNe reconstruction.

%
%
%
%
%
%
%

\section{AFBP Reconstruction Algorithm}\label{sect:AFBP}

We describe a design of a linear reconstruction transform, introduced
earlier in the conference publication \cite{ShEl09}.

\subsection{Definition of the AFBP Transform}\label{sect:AFBP_def}
\begin{figure}[htbp]
\vspace{0.3in} \centering

\begin{picture}(9.4,0.8)(0,0)
\put(0,0.4){\vector(1,0){1}} \put(0.5,0.6){\makebox(0,0)[b]{$
g_{\theta}(s) $}}

\put(1,0){\framebox(1.8,0.8){$ \mathbf{F}_{\kappa^{\mc{P}}} $}}
\put(1.9,1){\makebox(0,0)[b]{$\substack{Sinogram \\ filter}$}}
\put(2.8,0.4){\vector(1,0){1}}

\put(3.8,0){\framebox(1.8,0.8){$ \mathbf{R}^* $}}
\put(4.7,1){\makebox(0,0)[b]{$\substack{Back-\\ Projection}$}}
\put(5.6,0.4){\vector(1,0){1}}

\put(6.6,0){\framebox(1.8,0.8){$ \mathbf{F}_{\kappa^{\mc{A}}}  $}}
\put(7.5,1){\makebox(0,0)[b]{$\substack{Image \\ filter}$}}
\put(8.4,0.4){\vector(1,0){1}} \put(8.9,0.6){\makebox(0,0)[b]{$
\tilde{f}(x) $}}
\end{picture}
\caption{AFBP reconstruction scheme.\label{fig:AFBPscheme}}
\end{figure}
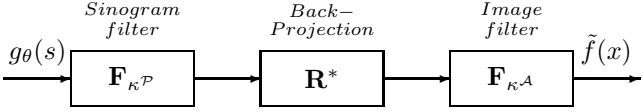

The extended linear reconstruction operator, labeled as {\em Adaptive
Filtered Back-Projection} (AFBP), is defined in the block diagram given in Figure
\ref{fig:AFBPscheme}, with parameter set $\kappa=\{\kappa^{\mc{P}},\kappa^{\mc{A}}\}$.

The linear filter $\mathbf{F}_{\kappa^{\mc{P}}}$ acts on a sinogram by
a distance\footnote{we refer here to the distance $s$ of a projection
bin from the center of the projection}-dependent 2-D convolution
kernel, which definition is detailed below.
The filter $\mathbf{F}_{\kappa^{\mc{A}}}$ in image domain is
implemented as a single 2-D convolution kernel applied to the output
of the Back-projection transform.The involved
kernels are generated in the training process,
described in the Section\ref{sect:AFBP_train}.

\textbf{Filter $\mathbf{F}_{\kappa^{\mc{P}}}$ - using 2-D convolution kernels:} The ideal inverse Radon operator
only requires a 1-D filter applied to each projection. In the
practical setup, we use a two-dimensional kernel applied to the
sinogram (each projection is also affected by the few neighboring
ones). Such a filter exploits the correlation between the
neighboring projections and can improve the reconstruction quality.

\begin{figure}[htbp]
\centering
\includegraphics[width=8cm]{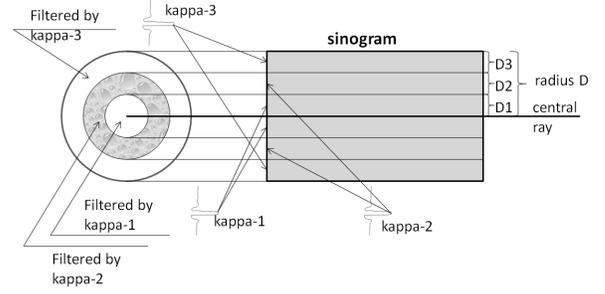}
\caption{Illustration of the radially-variant sinogram filtering. In
the left part of the diagram: concentric disks in image domain
corresponding to horizontal bands in the sinogram.}
\label{fig:img_diag1}
\end{figure}

\textbf{Filter $\mathbf{F}_{\kappa^{\mc{P}}}$ - using a distance-dependent kernel:} When the projections are
truncated to the ROI, the central part of each remaining projection
should be filtered using a symmetric kernel with a small spatial
support in order to reduce the truncation error. Near the edges the
information should be gathered in a non-symmetric way, only from the
non-truncated part of the projection. This can be done by assigning a
separate kernel to segments of radial distance from the center of the
projection. Explicitly, assume that only projections available are
over lines passing through a central disk of radius $D$ in the image domain.
We partition the range of distance $s \in [0,D]$ into $d$ disjoint
sub-segments $[0,D]=\bigcup_{i=1}^d D_i$, and use a bank of
$\kappa^{\mc{P}}=\{\kappa_i^{\mc{P}} \}_{i=1}^d$ of corresponding
convolution kernels. The filter is applied by convolving the two
projection segments corresponding to $\pm D_i$ with the kernel
$\kappa_i^{\mc{P}}$. See Figure \ref{fig:img_diag1}
for an illustration.

Here we use the AFBP algorithm in the setup of partial
data  - truncated projections. This technique also improves FBP performance
in the full-scan setup, but in this case the
gain is not substantial.


\subsection{AFBP Parameters Training}\label{sect:AFBP_train}
We state two goals pursued in the parameters training of the AFBP:

\textbf{Goal 1:} Assume availability of a representative set from some family of
images, X-ray intensity of the source, radius of a central disk
where the projections are measured, and a radius of
a central disk where the image should be reconstructed (ROI). The
goal is to maximize the reconstruction quality of $\mathbf{T}$, in
the Mean Square Error (MSE) sense, for these images (the
reconstruction is performed from noisy and truncated projections).

\textbf{Goal 2:} Under the same conditions but regardless the X-ray intensity,
design a reconstruction operator $\mathbf{T}$ with
such that the
action of the projection-reconstruction operator $\mathbf{TR}$ will be
as close as possible to the action of a radially-symmetric Gaussian
convolution kernel in image domain.

The second goal follows from the needs of the non-linear fusion
algorithm, which admits at its input a sequence of images
reconstructed in different ways. A special case of such sequence,
motivated by the LeGoNe denoising technique, is a sequence of images
with a growing measure of blur. The AFBP training, corresponding to
the stated second goal, will provide us with corresponding sequence
of linear operators.
\vskip0.3cm
\textbf{Objective function 1: best image quality.}
The objective function addressing the first
goal is designed as follows. Denote by $r$ the ROI radius. Let
$\mathbf{E}_{ROI}$ be the operator which nullifies all the bins in the
sinogram matrix which correspond to line integrals outside the central disk (in image domain) of radius $1.1r$. Also, we denote by
$\mathbf{F}_{ROI}$ an operator in the image domain that nullifies all
the image values outside the ROI.

Given a set $\mathcal{F}_{tr}$ of representative
images, we build the training set $\mathcal{G}_{tr}$ consisting of
noisy truncated sinograms
\begin{eqnarray}
\mathcal{G}_{tr}=\{ g^j_f =\mathbf{E}_{ROI}(\mathbf{R}f+\xi^j_f)\;|\;f \in
\mathcal{F}_{tr},\; j=1...J\},
\end{eqnarray}
where $\{g_f^j\}_{j=1}^J$ are obtained from $f \in \mathcal{F}_{tr}$
by applying the Radon transform and generating $J$ instances of
Poisson noise. Using this training set, the parameter set $\kappa$ of the AFBP transform is
then computed as an optimizer of the objective function
\begin{eqnarray}\label{eq:AFBP1}
\kappa^*=\arg\min_{\kappa}\sum_{f\in \mathcal{F}_{tr},j=1...J}\|
\mathbf{F}_{ROI}(\mathbf{T}_{\kappa} g_f^j-f)\|_2^{2},
\end{eqnarray}
where $\mathbf{T}_{\kappa}$ is defined in previous Subsection.

The objective (\ref{eq:AFBP1}) is quadratic in parameter sets $\kappa^{\mc{P}}$ and in $\kappa^{\mc{A}}$ separately. Thus, when
fixing one, the other is updated by solving a corresponding
linear problem (we use the Conjugate Gradients (CG) method). The training is then carried out in turns, fixing one set of variables and updating the other. We do not provide a convergence proof, but the property of the CG algorithm guarantees
a monotonous decrease of the objective function. In practice, we
continue the update process until the first $5$ significant digits of the objective function are stabilized. In our experience, no noticeable change in the algorithm behavior occurs in case of further optimization.

The resulting linear reconstruction transform $\mathbf{T}_{\kappa}$
has the following advantages over the conventional FBP:

\textbf{(1)} The objective function adapts the reconstruction operator to
partial data conditions, and thanks to its distance-dependent filter
such adaptation results in more adequate treatment of the truncated projections.

\textbf{(2)} The adjustment of the low-pas filter parameters in FBP is
replaced with automatic derivation of the filter, which cut-off
frequency adjusts to the typical spectra of training images and the noise
 intensity.

\textbf{(3)} The post-processing filter $\mathbf{F}_{\kappa^{\mc{A}}}$ is
matched to the projections filter. It helps reducing the value of
objective function substantially (as compared to using only
projections filter) and therefore to improve the output quality.

\textbf{(4)} The specialization of the reconstruction operator to the given
set of images makes it less universal, but improves the
reconstruction of similar images. By using a set of images typical
to the given task (for instance, sections of CT scans of specific
body part), a dedicated reconstruction transforms can be trained for
different clinical cases. Another application is to devote a
personal reconstructor to an individual patient and use her previous
scan data (say, from older similar CT scans) to allow lower X-ray
dosage in future scans.
\vskip0.3cm
\textbf{Objective function 2: approximate convolution filter.}
The objective function built for the second goal involves
noiseless projection sets. It has one additional parameter
- the standard deviation $\sigma$ of the
Gaussian kernel which action $\mathbf{TR}$ should mimic. The
parameter set $\kappa$ corresponding to the value of $\sigma$ is
computed as the optimizer of the following equation:

\begin{eqnarray}\label{eq:AFBP2}
\kappa^*=\arg\min_{\kappa}\sum_{f\in \mathcal{F}_{tr}}\|
(\mathbf{T}_{\kappa} \mathbf{R}f-\mathbf{F}_{ROI}(\mt{G}_{\sigma}*f))\|_2^{2},
\end{eqnarray}
This objective function encourages the reconstruction transform
$\mathbf{T}$ to produce a blurred output image, which resembles as
much as possible the "true" scanned image, filtered with the
corresponding Gaussian. Such reconstruction is an alternative to
using an FBP algorithm with a low-pass filter applied on the
projections. Advantage of the AFBP transform consists in the points
mentioned earlier and allows us to use the non-linear fusion in the
setup of ROI reconstruction from truncated projections.

\begin{figure}[htbp]
\centering \subfloat[Kernel for sharp
PSF]{\includegraphics[width=4.5cm]{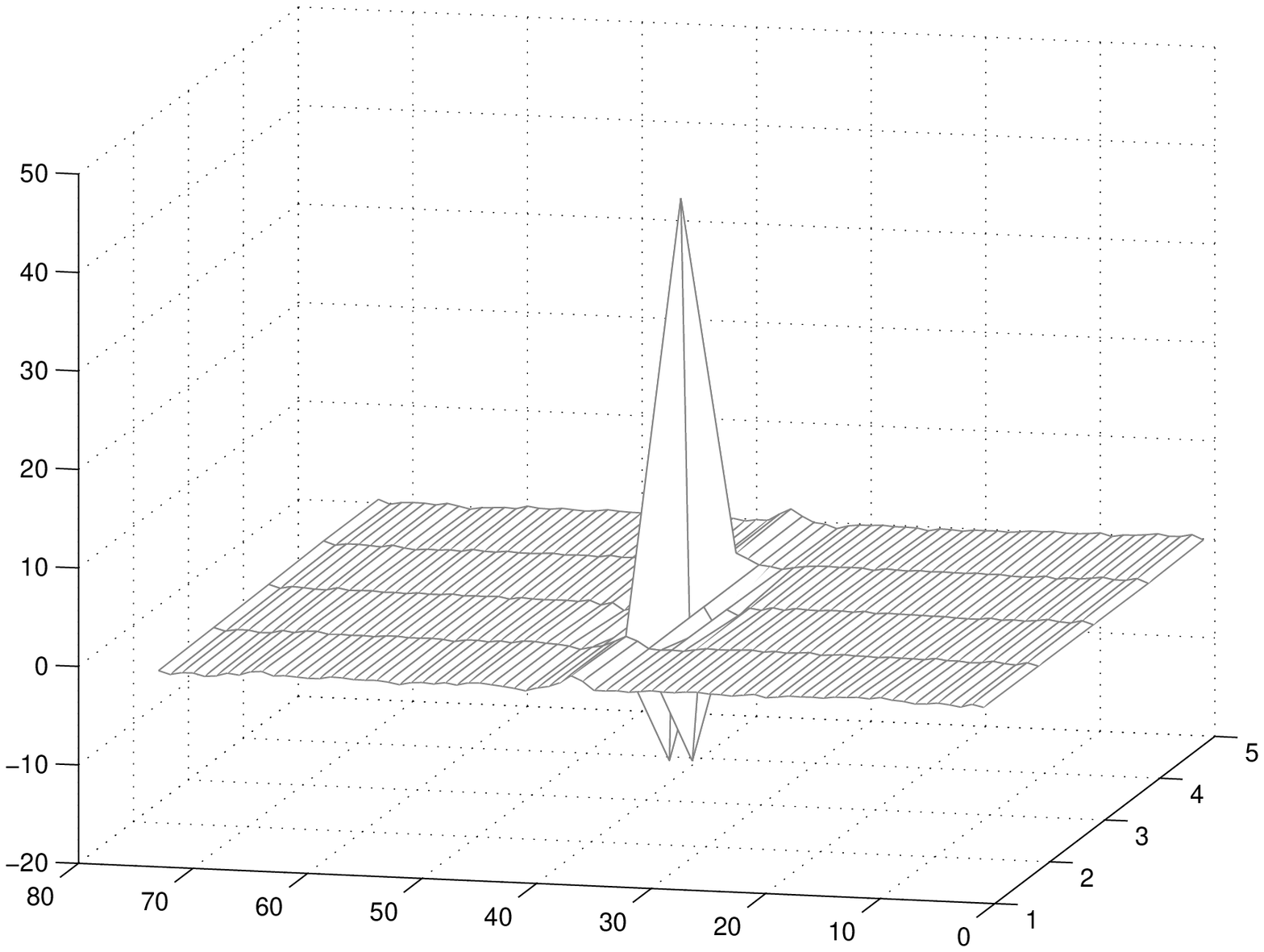}}
\subfloat[Kernel for wide-spread PSF]{\includegraphics[width=4.5cm]{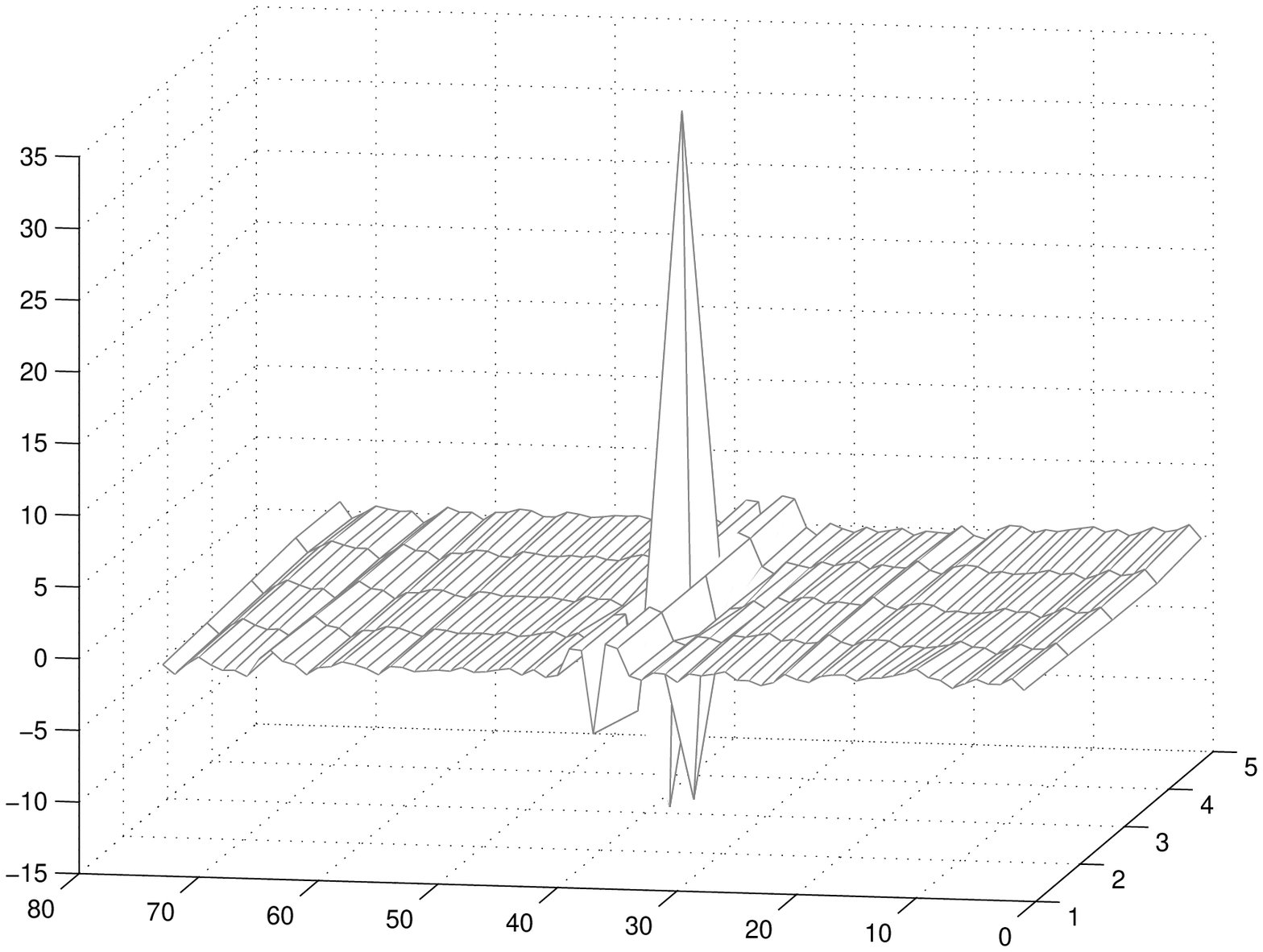}}\\
\centering \subfloat[Ram-Lak
kernel]{\includegraphics[width=3cm]{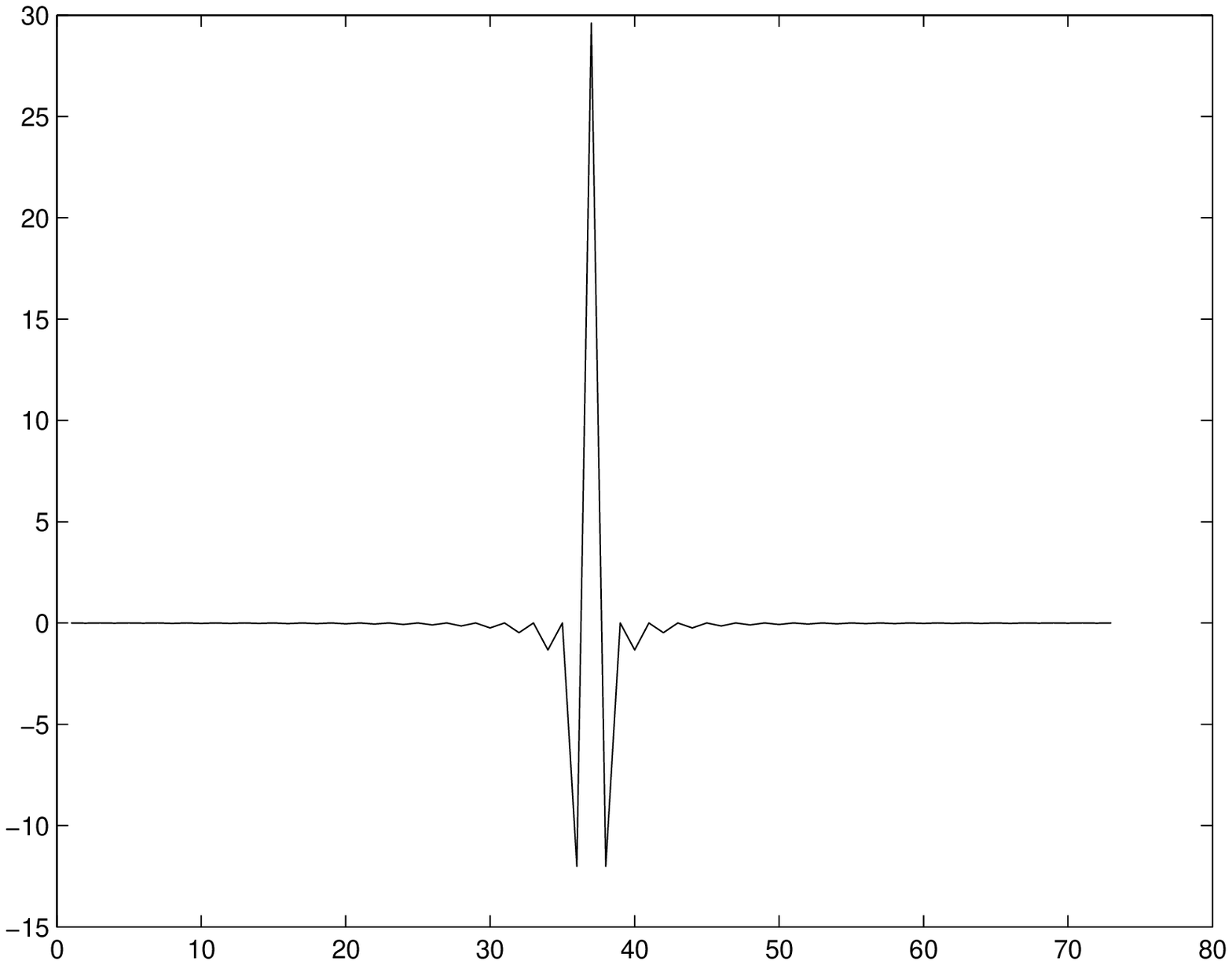}}
\caption{Bin values of the trained kernels, compared to the 1-D
Ram-Lak kernel.} \label{fig:ker_mesh}
\end{figure}

\textbf{Trained kernels: } We display visual examples of traned convolution kernels for the sinogram filter of AFBP.
in Figure \ref{fig:ker_mesh} displays two instances of such kernels, one
trained for sharp PSF of the $\mathbf{TR}$ operator, and the other one for a
wide-spread PSF. The central column of each
kernel resemble the Ram-Lak filter, and the neighbor columns ( corresponding to neighbor projections) are similar except for the central main peak.The bin values of kernel producing sharp PSF decay rapidly away from
the center of the kernel, and the bin values of the blurring kernel
are non-decaying, providing input from far projection bins in the
convolution process.

\subsection{Comparison to Previous Work}\label{sect:AnSa}
We return to the previously mentioned work reported in \cite{AnSa02}.
The Nonlinear Back-Projection (NBP) algorithm, proposed by its authors,
employs a different kind of a non-linear filter. It is
applied locally to the sinogram data, pre-filtered with the apodized
Ram-Lak filter, in the FBP framework. Each small 2-D sinogram patch is filtered
by a data-dependent combination of pre-learned filters. The parameters of these filters are statistically trained to yield the lowest MSE values in the image
domain for the training set of images. The proposed approach elegantly employs the Gaussian Mixture model in the sinogram domain and its patch-wise classification of the sinogram data has a very good potential to incorporate the sinogram features that would allow high-quality reconstruction.

This technique shares it conceptual approach with the AFBP and the SPADES
algorithms, proposed in this article, but can not be directly
compared to either one. AFBP also employs a spatially-variant,
statistically trained 2-D filter in the sinogram domain, but this
filter is linear and not data dependent.
SPADES is more closely related to NBP since both algorithms are locally adaptive:
NBP employs a sort of data-dependent fusion in the sinogram space, by
means of a combination of local filters, and SPADES acts in the image
domain, fusing the pre-computed linearly reconstructed images with a
neural network. It is difficult to argue in favor of one or the other
approach, except maybe for the fact SPADES uses weaker assumptions on
the reconstruction problem (we do not assume Gaussian mixture model
on small sinogram windows). Unfortunately, we weren't able to conduct a numerical comparison of the two techniques: the results on geometrical phantoms, presented in \cite{AnSa02}, are inferior to FBP, and the successful experiment on the clinical SPECT data can not be repeated for technical reasons.

\subsection{Numerical Experiments: AFBP on ROI Data}\label{sect:AFBP1-ROI}
We present numerical experiments of ROI
reconstruction from truncated projections. The first one is
performed with  geometric phantoms, described
in Section \ref{sect:LeGoNe}. The projections of $15$ training images
were truncated to a central disk of radius $32+3$
pixels. A Poisson noise, expressed in restriction of the source
intensity to $I_0 = 1200$, was applied to the data. The ROI consists
of a central disk of radius $32$ pixels. The AFBP projections filter
is a 2-D radially-variant convolution kernel, as
described in Section \ref{sect:AFBP_def}. Specifically, the size of each 2-D
kernel is $5\times 72$ [angles$\times$projection bins], and the ROI
radius is divided into $5$ sub-segments $D_i$ (overall, $5$ 2-D kernels).
The parameters of the AFBP algorithm were trained using equation
(\ref{eq:AFBP1}) and then applied to the truncated, noisy projections of
the test image.

\begin{figure}[htbp]
\centering
\includegraphics[width=7.1cm]{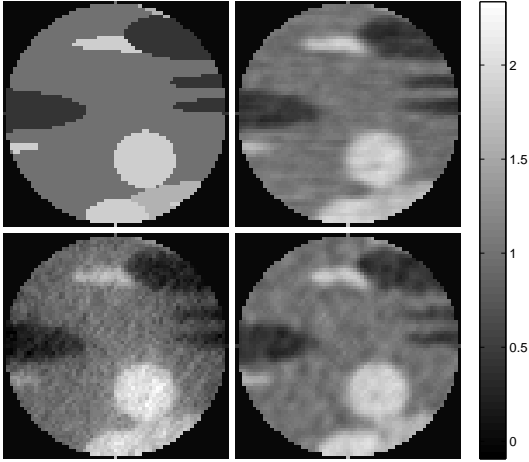}
\caption{Reconstruction in a ROI. Upper left: true image,  upper
right: AFBP reconstruction from truncated projections (SNR = 18.44
dB). Lower left: FBP reconstruction from truncated projections  (SNR
= 14.55 dB). Lower right: FBP reconstruction from full data (SNR =
18.06 dB).} \label{fig:AFBP-ROI1}
\end{figure}

\begin{figure}[htbp]
\centering
\includegraphics[width=7.1cm]{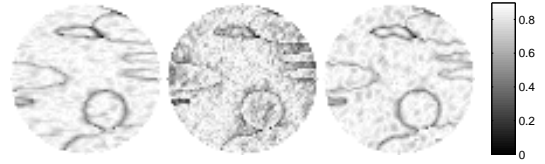}
\caption{Error images $|f_{orig}-\tilde{f}|$ in the ROI. Left to
right: AFBP on truncated projections,  FBP
on truncated projections,  FBP on full data. Darker shades correspond to larger error.}
\label{fig:AFBP-ROI2}
\end{figure}

\begin{figure}[hpbp]
\centering
\includegraphics[width=4cm]{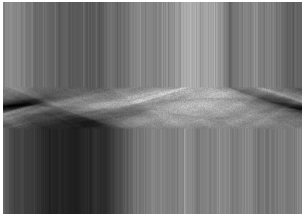}
\caption{Sinogram completion method used for FBP. The horizontal
band in the middle of the image represent the available projections,
truncated to the central disc in image domain. Other rows are
replica of the upper and lower margins of the band. This simple step
helps improving the FBP performance substantially.}
\label{fig:img_9a}
\end{figure}

For comparison we use the FBP algorithm which is implemented as
follows: (1) The Ram-Lak kernel is smoothed by applying a
Butterworth apodization window in the frequency domain. The parameters of the window function are tuned on the training set to provide the best SNR value. (2) The truncated projections are extrapolated by replicating the first and the last
non-zero rays (see Figure \ref{fig:img_9a}). This simple but effective linear
sinogram completion technique was described in \cite{FaWa97}. See Figures \ref{fig:AFBP-ROI1},\ref{fig:AFBP-ROI2} for a visual comparison of the algorithms.

%
%

\begin{figure}[htbp]
\centering
\includegraphics[width=8.1cm]{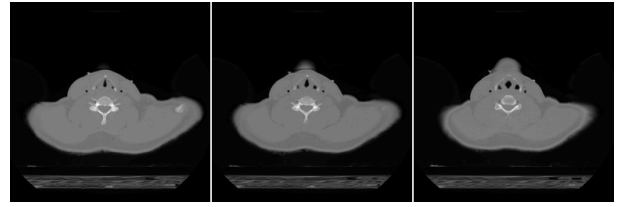}
\caption{Clinical CT images from the AFBPt training set}
\label{fig:VHhead_trset}
\end{figure}
\begin{figure}[htbp]
\centering
\includegraphics[width=6.4cm]{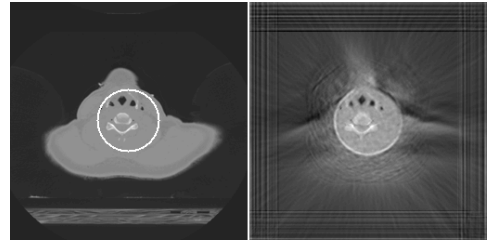}
\caption{Left: true image with a marked Region Of Interest. Right:
full reconstruction from truncated projections by AFBP. The ROI is
clearly distinguished.} \label{fig:AFBPt_ROI1}
\end{figure}
\begin{figure}[htbp]
\centering
\includegraphics[width=6.1cm]{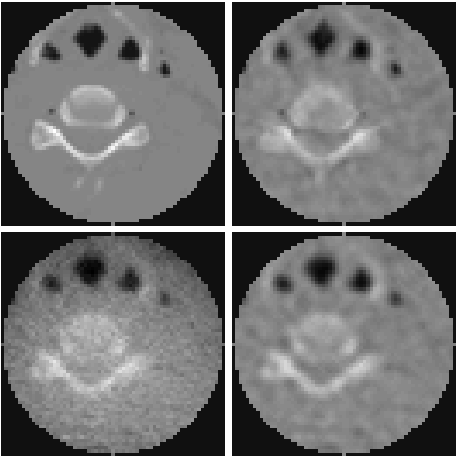}
\caption{Reconstruction in a ROI. Upper left: true image,  upper
right: AFBP on truncated projections (SNR = 22.05
dB). Lower left: FBP on truncated projections  (SNR
= 15.73 dB). Lower right: FBP on full data (SNR =
22.28 dB).} \label{fig:AFBPt_ROI2}
\end{figure}

The experiment was repeated on a set of clinical CT images, which represent axial head sections (courtesy of the \textit{Visible Human}
project\footnote{\text{www.nlm.nih.gov/research/visible/visible\_human.html}})
The CT scan was performed on a GE scanner and consists originally of
$512\times 512$ grayscale images with pixel depth of 12 bits. The
images were cropped and resized to $256\times 256$ size for the purpose of our experiments. Some of them are displayed at Figure \ref{fig:VHhead_trset}). Figure
\ref{fig:AFBPt_ROI1} displays the ROI and a typical result of reconstruction
from truncated projections. In Figure \ref{fig:AFBPt_ROI2} we compare
the AFBP versus the FBP algorithm.

It can be seen from the images that AFBP improves substantially the FBP output,
producing an image comparable to a reconstruction from full scan data.
The SNR values support this observation.

\subsection{Study of AFBP and FBP Algorithms on ROI}

We now turn to second kind of AFBP transforms $\mathbf{T}_i$, trained by optimizing the equation (\ref{eq:AFBP2}) for different values of $\sigma_i$. These reconstruction transforms come to replace a sequence of FBP transforms $\{\mathbf{S}_i\}_{i=1}^I$
in ROI reconstruction, due to the need to adjust the linear
reconstructors to the missing data setup: FBP is theoretically
derived for full-data scan and its performance of truncated
projections is poor, even with optimized parameters of the low-pass
window\footnote{which indeed make a significant difference in FBP
performance.}.

Now we define a measure on a reconstruction transform $\mathbf{X}$
that expresses the amount of blur $\mathbf{X}$ introduces into the output
image. Let $\mt{G}_{\sigma}$ stand for the rotationally symmetric 2-D
Gaussian convolution kernel with standard deviation $\sigma$. Given a training set $\mathcal{F}_{tr}=\{f_1,f_2,...,f_N\}$, we define
\begin{equation}\label{eq:Gauss_sigma}
\zeta_{\mathbf{X}}= arg\min_{\sigma}=\frac{1}{N}\sum_{j =
1}^{N}\|\mathbf{XR}f_j-\mt{G}_{\sigma}*f_j\|_2
\end{equation}
In words, the {\em blur measure} $\zeta_{\mathbf{X}}$ is the standard deviation of a Gaussian kernel which action on images of the training set resembles the most (in MSE sense) the action of $\mathbf{XR}$. Notice that the value of $\zeta_{\mathbf{X}}$ is, up to a constant, the Full-Width-Half-Maximum (FWHM) of the effective Gaussian blurring kernel, since the FWHM of a Gaussian is $2.35$ multiplied by its standard deviation.

Notice that for the AFBP transform $\mathbf{T}$ trained via equation
(\ref{eq:AFBP2}) with corresponding standard deviation $\sigma_T$, we
have $\zeta_{\mathbf{T}} = \sigma_T$. For any other transform
$\mathbf{X}$ the blur measure can be easily approximated by
evaluating the optimizer of (\ref{eq:Gauss_sigma}) via a grid search
on values of $\sigma$.

The blur measure will allow us to parameterize the sequences $\{\mathbf{T}_i\},\{\mathbf{S}_i\}$
of reconstruction algorithms in order to correctly compare their performance: for a given level of noise present in the sinogram, the output quality is a function of the blur measure applied in the reconstruction process. More importantly, this notion allows us to unambiguously characterize the sequence of reconstructors, employed in the first stage of the SPADES algorithm, instead of specifying the parameters of apodization window for FBP or the values of $\sigma_i$ for AFBP.

\begin{figure}[htbp]
\centering
\includegraphics[width=6cm]{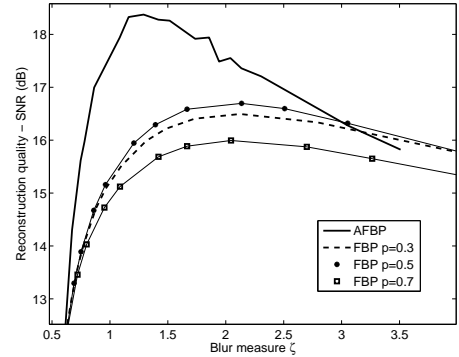}
\caption{Graphs of reconstruction quality versus the blur measure
$\zeta_{\mathbf{T}}$, computed for the sequence of AFBP transforms and
also for three sequences of apodized FBP transforms with different
degrees $p$ of Butterworth window. } \label{fig:FWHM}
\end{figure}

In the following numerical experiment, a sequence $\{\mathbf{T}_i\}_i$ was trained on $15$ geometric phantoms by optimizing the equation\ref{eq:AFBP2},
in the setup described in the beginning of the Section \ref{sect:AFBP1-ROI}.
Using these phantoms, the values $\zeta_{\mathbf{T_i}}$ are computed
empirically, as well as the values of  $\zeta_{\mathbf{S}_j}$ for FBP
transforms with corresponding cut-off frequency $q_j$ (see equation (\ref{eq:BW})). We used different values of the parameter $p$, to achieve best available quality for the FBP. All algorithms are
applied to noisy truncated projections, mended by the aforementioned completion technique (see Figure
\ref{fig:img_9a}). In Figure \ref{fig:FWHM} we present graphs of SNR
values of phantoms reconstructed from noisy and truncated
projections, with the same setup as in the previous experiment.
Specifically, we plot average SNR values over a set of phantoms, as
function of the blur measure. The comparison is carried out between
the AFBP and the apodized FBP with different values of $p$.
It can be seen that with $p=0.5$, FBP achieves maximal quality at
$\zeta \sim 2.1$, which is substantially lower than SNR values
achieved by the trained kernel of AFBP at $\zeta\sim 1.2$.

%
%
%
%
%
%
%

\section{SPADES in the ROI}\label{sect:SPADES_AFBP}
\subsection{The Algorithm}
We return to the SPADES algorithm (see Section
\ref{sect:SPADES_FBP}) with AFBP transforms. Now it can be extended
to the setup of ROI reconstruction from
truncated projections. To that end, AFBP transforms of two types,
corresponding to objective functions (\ref{eq:AFBP1}),(\ref{eq:AFBP2}) are used:

\noindent\textbf{(1)} One
transform $\mathbf{T}_o$, trained with (\ref{eq:AFBP1}), will
produce an image of best quality (in SNR terms) attainable with AFBP
at the given noise level, for images similar to those of the
training set.
\noindent\textbf{(2)} A sequence of transforms $\{\mathbf{T}_i\}_{i=1}^I$
trained via the equation (\ref{eq:AFBP2}),
will produce a set of image versions with a growing blur, corresponding to
 standard deviation values
$\{\sigma_i\}_{i=1}^I\}$. we use $I=10$ such transforms.

The training of AFBP versions is performed on a given set of images.
Afterwards, the same training set is used to compute the weights of
the neural network, while the inputs are obtained from the linear
AFBP image estimates. When all components of the SPADES
reconstruction are ready, the processing of the sinogram data $g_n$
for an unknown image is carried out in three steps (see a diagram in
the Figure \ref{fig:SPADES_scheme}):
\begin{enumerate}
\item
Compute preliminary images $\tilde{f}_i=\mathbf{T}_ig_n,
i=1,2,...,I$, and $\tilde{f}_o = \mathbf{T}_o g_n$.
\item
For each pixel $p$ in the ROI, build a vector $x_p$ of features from the
corresponding values of $\tilde{f}_i,\tilde{f}_o$ (see Section
\ref{sect:NN} for details).
\item
Use the neural network to compute the value of the output image at
$p$ from the vector of features $x_p$.
\end{enumerate}

\subsection{Computational Complexity of SPADES}
The computational complexity is calculated for images of size $n\times n$ and for $I+1$ preliminary reconstruction operators.

\textbf{(1)} The standard FBP algorithm, which consists of a 1-D filtering of the sinogram columns and a Back-Projection transform, requires
$\mathbb{O}(n^3)$ computations for each of these steps. The AFBP
differs in the $2-D$ sinogram filter, which uses about $3-5$ times more operations than a $1-D$ filter, and in a 2-D post-processing filter which consumes $\mathbb{O}(n^3)$ computations: we used a $2-D$ kernel of dimensions $\sqrt{n}\times\sqrt{n}$, therefore its action requires $n$ operations per pixel. Therefore AFBP has the overall complexity of $\mathbb{O}(n^3)$ computations.

\textbf{(2)} SPADES involves $I+1$ versions of the AFBP transform.  If the
computing machinery allows to execute the $I+1$ AFBP algorithms in
parallel, time consumption of the preliminary stage equals to a
single $FBP$ reconstruction. Otherwise, the preliminary step
requires $\mathbb{O}(I\cdot n^3)$ operations. Typical value of $I$,
used in our experiments, is $I=10$.

\textbf{(3)} The neural network involves $(I+11)N$ weights, which participate in the computation of every pixel in the image. Here $N$ is a number of neurons and
(again, in our experieece) it should be roughly equal to the size of
the vector of features, $I+10$. Thus, computation of each intensity
value for the output image requires $\sim\ (I+10)^2\sim 400$
operations.

\textbf{(4)} If the number of pixels in ROI is $M$, the overall
complexity is $\mathbb{O}((I+1)n^3+(I+10)^2M)\sim\mathbb{O}(10n^3+400M)$
operations.

\subsection{Numerical Experiments - SPADES on the ROI data}
In the first numerical experiment, we used a training set of $15$
geometric phantoms, described earlier. The chosen blur measure values
$\zeta_{\mathbf{T}_i}$ are equally spaced in the interval $[0,3.5]$.
Linear reconstruction of the ROI in a test image with AFBP algorithms
was conducted from noisy and truncated projections (see experiment setup in Section \ref{sect:AFBP1-ROI}. The neural network was applied on the resulting images to produce the final output.

\begin{figure}[htbp]
\centering
\includegraphics[width=7.1cm]{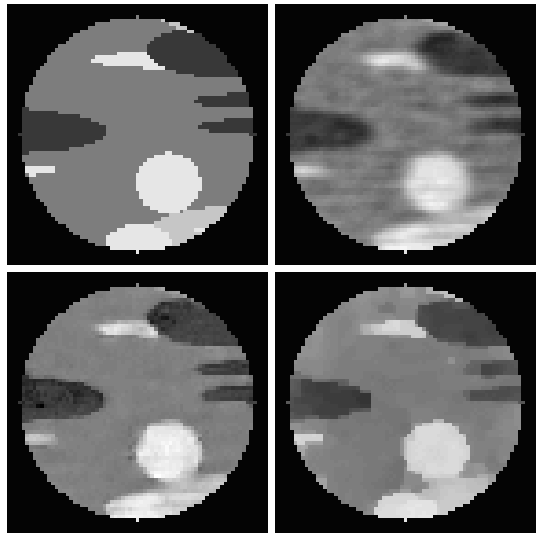}
\caption{Upper left: true image. Upper right:  best AFBP version
(18.44 dB). Lower left: SPADES reconstruction (19.81 dB). Lower
left: Statistical reconstruction (19 dB).} \label{fig:SPADES_results1}
\end{figure}

Outcomes of three different algorithms are visually compared in Figure
\ref{fig:SPADES_results1}: a linear reconstruction scheme AFBP, a
locally-adaptive direct algorithm SPADES and a statistically based
iterative algorithm \cite{ElFe02} (see the
Appendix section for details of our implementation). It can be
observed that SPADES has succeeded in reducing the background noise,
stemming from low X-ray intensity, restoring the piece-wise constant
texture of the phantom. The SNR value of SPADES output is higher by
$1.15$ dB than the SNR of best linear algorithm output we could
produce. The statistical reconstruction result has sharper edges and
more homogeneous ellipses, at the price of higher computational time.
Notice that because of artifacts its SNR value is still lower than
that of SPADES output. The reconstruction was further carried out on $23$ other test images.
The average SNR values of the ROI quality are: FBP (on truncated projections) - 15.80 dB,
FBP (on full data) - 18.89 dB, AFBP (on truncated projections) - 19.31 dB, SPADES (on truncated projections) - 20.30 dB. Such quality improvement was possible due to the fact FBP displays discretization artifacts, which restrict the reconstruction quality even in the absence of noise and with no truncation of the projections. The learned kernels of AFBP succeed to partially compensate this drawback.

The experiment was repeated for a ROI reconstruction in the Forbild
head phantom, designed in the framework of the Forbild project
\footnote{http://www.imp.uni-erlangen.de/forbild/english/results/head/head.html.}.
A set of ten similar phantoms, obtained from the head phantom by
random perturbations of the radii and locations of the composing
ellipses, served as a training set for SPADES. The behavior of the
involved algorithms (AFBP, SPADES and the statistical reconstruction)
was similar to one observed in the Figure \ref{fig:SPADES_results1}.
Numerically, the quality of ROI recovered by AFBP in the Forbild phantom was 17.93 dB, statistical reconstruction has achieved the same value, and SPADES
has demonstrated 20.1 dB. Due to the lack of space, we omit the
accompanying graphics.

\begin{figure}[htbp]
\centering
\includegraphics[width=7.1cm]{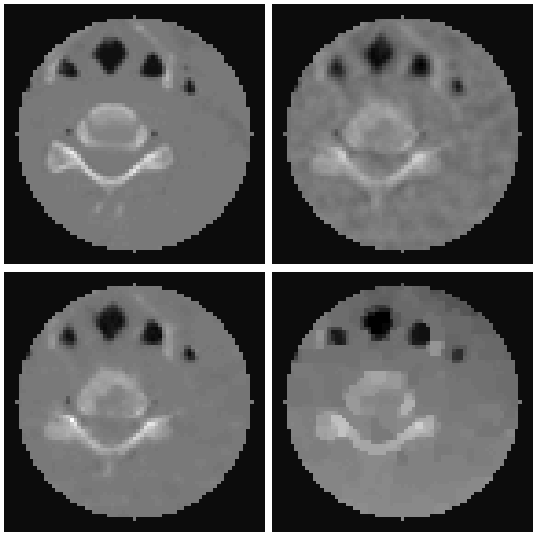}
\caption{Upper left: true image. Upper right:  best linear AFBP
(22.04 dB). Lower left: SPADES reconstruction (24.58 dB). Lower
left: Statistical reconstruction (19.24 dB)} \label{fig:SPADES_results3}
\end{figure}

Another experiment was carried out with the clinical CT
images, described earlier in the Section \ref{sect:AFBP1-ROI}. The
AFBP output, generated in the earlier experiment on CT data (Section
\ref{sect:AFBP1-ROI}, Figure \ref{fig:AFBPt_ROI2}), is used, along
with a sequence of blurred versions of the test image, as the input
data for the neural network. The parameters of AFBP , as well as the
neural network, were trained using a set of $10$ clinical images
(some are presented in Figure \ref{fig:VHhead_trset}). Output of
AFBP, SPADES and the statistical algorithm are displayed in Figure
\ref{fig:SPADES_results3}. Their behavior is
similar to the one observed on the phantoms. The linear AFBP produced
a good\footnote{see section \ref{sect:AFBP1-ROI} for comparison with
FBP} preliminary image, which nevertheless contains noise propagated
from the projections. SPADES succeeds to remove the noise
substantially, resulting in an image close to the ground truth. The
statistical algorithm also produces a clean image, but the artifacts
introduced by the Huber prior deteriorate its quality. SNR values
reflect on these observations.

In the reconstruction of $16$ additional test images, average SNR gain of the SPADES algorithm over the AFBP was $1.93$ dB. This shows consistency in the behavior of the two algorithms, visually displayed in the Figure \ref{fig:SPADES_results3}.

%
%
%
%
%
%
%

\section{Discussion and Conclusions}\label{sect:EPILOG}
We have proposed and demonstrated a CT reconstruction algorithm
based on a local fusion of linear estimates of the sought image. The
linear reconstructors were designed to automatically adapt to the
specific family of images and the noise level, as well as the
missing data setup. Resulting direct non-linear reconstruction
algorithm produces images inherently different from those obtained
with FBP or other linear methods: the noise level and artifacts are
substantially reduced, while the edges present in the image are well
preserved. We notice that the proposed method can in principle
improve any given reconstruction method, by fusing its output image
with a number of other image estimates by an adequately trained
neural network.

The proposed scheme involves very few parameters, mostly the design
details - number of linear reconstructors and their blur measure
values, structure of the vector of features and number of neurons.
All these are easily tuned in a practical setup of specific
implementation of the algorithm. Admittedly, the neural network
requires a delicate treatment, but in our experience (and especially
for specialists in Machine Learning) it is fairly manageable. In fact,
we used the simplest one-level feed-forward model of the neural network,
and it is plausible that using more complex network (with larger
training set) will improve the algorithm behavior.

We notice that the proposed scheme performs an isotropic processing
in the image domain: the inputs of the neural network are preliminary
images $\tilde{f}_i = \mathbf{TR}f$ which approximate the true image
$f(x)$ convolved with a {\em radially symmetric} kernel. The FBP
algorithm also has a radially symmetric PSF. A promising
future direction of work is an extension of SPADES to an anisotropic
algorithm, where elliptical kernels with varying directions and
dimensions are used. The neural network should automatically choose
the relevant direction (or combination of some), to use the
appropriate preliminary image version at each point

All the numerical results presented in this paper were obtained from
experiments executed entirely in the Matlab environment. The "ground
truth" images are projected using one specific implementation of the
discrete Radon transform, and also used as the ideal reference for
the training purposes. Admittedly, the results are affected by this
setup, hence the experiments should be rendered as
synthetic. The proposed methods can be extended to a practical setup
by using mechanical phantoms with geometric details and their
discrete computerized models. Despite the absence of realistic clinical data
experiments and industrial implementations of the FBP in the experiments, the differences observed between the existing and the
proposed techniques are consistent and explicit. Therefore, we hope
that presented techniques will also exhibit similar behavior when
applied in the industrial CT setup.

Notice that incorporating the proposed algorithm into an existing CT
machinery will only require minor software modifications -
replacement of FBP filters and addition of a learning machine. Since
SPADES is not slower than the standard FBP algorithm (when parallel
computation of FBP instances is available),
it represents an appealing alternative for full-scan reconstruction
and especially for ROI recovery from partial data.

In a broader view of numerical algorithms for inverse problems, the
proposed technique has quite a general underlying principle: a fusion
of a sequence of solutions with a certain structure, using a
discriminatively trained learning machine. One possible application
of this principle can be considered for the very common method of
solving an inverse problem by optimization of the objective function
of form
\begin{equation}
x^* = arg\min_{x}\{D(x,y)+\lambda P(x)\}.
\end{equation}
Here $y$ is a set of measurements, $x$ is the sought signal, $D(x,y)$
is the Data component which encourages the consistency of the
solution with the observations and $P(x)$ is a Prior component,
expressing the beliefs on the nature of the signal $x$. The optimizer
is usually obtained by an iterative update of the initial solution
$x_0$. The parameters of the algorithm, such as the value of $\lambda$
and the number of iterations, are crucial for successful
reconstruction but are difficult to estimate and often are
data-dependent.

An appealing solution to the problem of chosing the parameters would be to
compute a set of optimizers $x^*_i$, corresponding to different values
of $\lambda$ and/or the number of iterations. These preliminary
optimizers would then be fused into the final output by means of a
neural network (or other learning machine), trained on samples from a
training set. Such approach (on expense of higher computational
complexity) is expected to produce a solution $\tilde{x}$ of quality
higher than any individual $x^*_i$, as was the case in numerical
algorithms presented in this paper.

%
%
%
%
%
%
%

\appendix
\section{Appendix}
\subsection{Generating the Poisson Noise}
The Poisson noise is introduced into photon counts $y_i$ as a
consequence of limited values of $y_i$ themselves; the lower the
total count of the photons, received by a detector per unit of time,
the smaller the accuracy of the measurement for the corresponding
line integral. The mathematical model of such noise is the Poisson
distribution of photon
counts: $y_i$ is an instance of the random variable $Y_i$,
\begin{equation}
Y_i \sim Poisson(\lambda_i),\;\;\; \lambda_i = I_0e^{-(\mathbf{R}f)_i}.
\end{equation}
Here $f_0$ is the scanned image (a 2-D slice of an object), $I_0$ is
the photon count in absence of obstacles (proportional to the
initial X-ray intensity), and $(\mathbf{R}f)_i$ is the line integral
of the image along the line corresponding to detector $i$.

In our synthetic experiments, we first compute $\mathbf{R}f$ and
then generate the set of photon counts. This is done as follows: \textbf{(1)}
Denote $g = \mathbf{R}f$ and compute $y^0_i = I_0e^{-(g)_i}$ for every bin $i$ in the sinogram.
Naturally, the maximal photon count is $\max_i\{y^0_i\}=I_0$ since $min(g)=0$. The minimal count $y_{min} =\min_i\{y^0_i\}$ depends on the scaling of $g$. \textbf{(2)} We chose this scaling such that $y_{min}=60$, in order to avoid the problematic numerical behavior related to a low photon count, contaminated with the Poisson noise. Notice that the ratio $I_0/y_{min}$  should be constant when different values of $I_0$ are used.
\textbf{(3)} Draw instances of random variables $y^n_i \sim Poisson(y^0_i)$,
index $i$ runs over all the bins in the sinogram.
\textbf{(4)} convert back to the sinogram domain, by computing $g_n = -log(y^n/I_0)$.

\subsection{Statistically-based Iterative Reconstruction}
We implement a Penalized-Likelihood reconstruction algorithm, defined in \cite{ElFe02} for mono-energetic X-ray scan.
Given the noisy measurements $y$ of photon counts, the output image
$\tilde{f}$ is computed by minimizing the Penalized Likelihood
equation
\begin{equation}\label{eq:PL}
\Psi(f) = L(f|y)+\beta R_{\delta}(f)
\end{equation}
where $L(f|y)$ is the negative log-likelihood of the Poisson distribution,
\begin{equation}\label{eq:LL}
L(f|y) = \sum_{i=1}^N \left\{\lambda_i - y_i log(\lambda_i)+log(y_i!)\right\},
 \;\;\; \lambda_i = I_0 e^{-(\mathbf{R}f)_i}
\end{equation}
and $R_{\delta}(f)$ is an edge-preserving penalty which encourages

image smoothness:
\begin{equation}
R_{\delta}(f) = \sum_j\left\{\psi_{\delta}(\mathbf{D}f)_j\right\}
\end{equation}
with $\mathbf{D}f=[\nabla_xf,\nabla_yf]$ consisting of the directional derivative maps of the image $f$. We compute the
derivatives by the central difference approximation. It is defined
for $x$-axis by $f_x(px,py) = f(px+1,py)-f(px-1,py)$ and similarly for the $y$-axis. The Huber
penalty $\psi_{\delta}$ is defined by

\[ \psi_{\delta}(x) = \left\{ \begin{array}{ll}
         \dfrac{x^2}{2} & \mbox{if $|x|<\delta$},\\
        \delta|x|-\dfrac{\delta^2}{2}& \mbox{if $|x|\geq \delta$}.\end{array} \right. \]

The minimizer of equation (\ref{eq:PL}) is computed using a Gradient
Descent with a simple line-search function (we did not pursue a
computationally effective implementation, since only the quality of
the algorithm output is considered). The parameters $\beta,\delta$
and the number of iterations were tuned on the training images for each image to
attain the best possible SNR value.

\bibliographystyle{IEEEbib}
\bibliography{spaFBP_bib}

\begin{thebibliography}{10}

\bibitem{RaLa71}
G.~N. Ramachandran and A.~V. Lakshminarayanan,
\newblock ``{Three-dimensional Reconstruction from Radiographs and Electron
  Micrographs: Application of Convolutions instead of Fourier Transforms},''
\newblock {\em Proceedings of the National Academy of Sciences of the United
  States of America}, vol. 68, no. 9, pp. 2236--2240.

\bibitem{KiKi96}
K.I.~Kim J.H.~Kim and C.E. Kwark,
\newblock ``A filter design for optimization of lesion detection in spect,''
\newblock {\em IEEE Nuclear Science Symposium}, vol. 3, pp. 1683--1687, 1996.

\bibitem{OlDe94}
T.~Olson and J.~DeStefano,
\newblock ``Wavelet localization of the radon transform,''
\newblock {\em IEEE Trans Signal Process.}, vol. 42, no. 8, pp. 2055--2067,
  Aug. 1994.

\bibitem{DeBr95}
A.H. Delaney and Y.~Bresler,
\newblock ``Multiresolution tomographic reconstruction using wavelets,''
\newblock {\em IEEE Trans. on Image Processing}, vol. 4, no. 6, pp. 799 -- 813,
  Jun. 1995.

\bibitem{FaWa97}
C.A.~Berenstein F.~Rashid-Farrokhi, K.J. Ray~Liu and D.~Walnut,
\newblock ``Wavelet-based multiresolution local tomography,''
\newblock {\em IEEE Trans. Med. Imag.}, vol. 6, no. 10, pp. 1412--1430, Oct.
  1997.

\bibitem{ElFe02}
I.A. Elbakri and J.A. Fessler,
\newblock ``Statistical image reconstruction for polyenergetic x-ray computed
  tomography,''
\newblock {\em IEEE Trans. Med. Imag.}, vol. 21, no. 2, pp. 89--99, Feb. 2002.

\bibitem{Fess06}
J.A. Fessler,
\newblock ``Iterative methods for image reconstruction,''
\newblock in {\em IEEE International Symposium on Biomedical Imaging}.

\bibitem{AnSa02}
K.D.~Sauer B.I.~Andia and C.A. Bouman,
\newblock ``Nonlinear backprojection for tomographic reconstruction,''
\newblock {\em IEEE Trans Nucl Sci}, vol. 49, no. 1, pp. 61--68, Feb. 2002.

\bibitem{ShEl08}
M.~Zibulevsky J.~Shtok, M.~Elad,
\newblock ``Adaptive filtered-back-projection for computed tomography,''
\newblock in {\em IEEE 25-Th Convention Of Electrical And Electronics Engineers
  In Israel}. Mar. 2008, pp. 528--532, IEEE.

\bibitem{ShEl09}
M.~Zibulevsky J.~Shtok, M.~Elad,
\newblock ``Direct adaptive reconstruction algorithms for computed
  tomography,''
\newblock in {\em IEEE International Symposium on Biomedical Imaging: From Nano
  to Macro}, 2009.

\bibitem{GoNe97}
A.~Goldenshluger and A.~Nemirovsky,
\newblock ``On spatial adaptive estimation of nonparametric regression,''
\newblock {\em Math. Meth. Statist.}, vol. 6, no. 2, pp. 135--–170, 1997.

\bibitem{RiBi06}
J.~Bian P.J. La~Rivière and P.A. Vargas,
\newblock ``Penalized-likelihood sinogram restoration for computed
  tomography,''
\newblock {\em IEEE Trans. Med. Imag.}, vol. 25, no. 8, pp. 1022--36, Aug.
  2006.

\bibitem{LeMa97}
V.G.~Spokoiny O.V.~Lepskii, E.~Mammen,
\newblock ``Optimal spatial adaptation to inhomogeneous smoothness: An approach
  based on kernel estimates with variable bandwidth selectors,''
\newblock {\em Ann. Statist.}, vol. 25, no. 3, pp. 929--947, 1997.

\bibitem{Leps90}
O.V. Lepski,
\newblock ``One problem of adaptive estimation in gaussian white noise,''
\newblock {\em Theory Probab. Appl.}, vol. 35, no. 3, pp. 459--470, 1990.

\end{thebibliography}

\end{document}